\documentclass[11pt]{article}
\usepackage[margin=1in]{geometry}
\usepackage{amsmath,amssymb}
\usepackage{booktabs}
\usepackage{multirow}
\usepackage{hyperref}
\usepackage{microtype}
\usepackage{listings}
\usepackage{xcolor}
\usepackage{graphicx}
\usepackage[T1]{fontenc}
\usepackage[numbers,sort&compress]{natbib}

\title{Mechanism-level routing failure in LLMs\\
       over Lean-verified algebraic structures}

\author{
  Manuel Israel C\'azares\textsuperscript{1} \and
  Wenlin Zhang\textsuperscript{2} \and
  Haobo Ma\textsuperscript{3}
}

\date{
  \textsuperscript{1}Bytepro AI, Mazatl\'an, M\'exico\\
  \textsuperscript{2}National University of Singapore / Omega Institute\\
  \textsuperscript{3}ChronoAI / Omega Institute\\[6pt]
  \small\textit{Correspondence:} hello@bytepro.ai, e1327962@u.nus.edu, aloning@gmail.com
}

\begin{document}
\maketitle

\begin{abstract}
We present an empirical study of structural routing failure in large language
models (LLMs) over a formally verified algebraic corpus. The task requires
selecting the correct proof-mechanism label---from a fixed closed
template set---for compact mathematical objects drawn from the FiberRing
formalization in Lean~4~\cite{omega2026automath}, where each evaluation item
is anchored to a Lean-verified artifact and assigned a proof-mechanism label
from the corresponding named certificate family.

Our central finding is a \textbf{mechanism-level routing ceiling}: under
blind conditions (no structural cue), gpt-oss-120b achieves $80.3\%$
template accuracy on 22 clean-anchor FiberRing items
($n=66$ evaluations; temperature~$=0$, seed~$=0$, reasoning~$=\mathrm{low}$),
while Llama~3.3~70B reaches $68.2\%$.
Exposing a mechanism-bearing Lean verdict / witness cue drawn from the
formalization (Condition~A2) raises accuracy to $90.9\%$ and $81.8\%$
respectively~---~a gap of $+10.6$ and $+13.6$ percentage points that we term
\textbf{cue-induced routing uplift}.

The dominant failure mode is a CRT\,$\to$\,ring-equivalence misroute:
gpt-oss-120b misroutes 7 of 12 CRT items ($58.3\%$) in the blind condition,
and zero in the cue-conditioned condition. We show this error has a direct
correlate in the Lean corpus: every CRT row is presented as a ring
equivalence at the outer type level, while the finer mechanism is the coprime
Fibonacci-modulus factorization followed by \texttt{ZMod.chineseRemainder}.
The model identifies the broad
algebraic type but fails to resolve the finer proof mechanism without the
cue.

Two \textbf{cue-resistant} failure modes persist in both conditions
across both models:
\begin{itemize}
  \item \texttt{stable{-}value{-}map} $\to$ \texttt{ring{-}structure}
  \item \texttt{prime{-}power{-}fiber{-}specialization} $\to$
        \texttt{ring{-}equivalence}
\end{itemize}
These errors identify two different boundary phenomena in the Lean
formalization. The \texttt{stable{-}value{-}map}$\,{\to}\,$\texttt{ring{-}structure}
error is a clean formal distinction with strong surface-language overlap:
map-preservation rows mention addition, multiplication, zero, one, and ring
homomorphism, but their proof role is transport through \texttt{toZMod}, not
construction of the ring instance on $X_m$. The
\texttt{prime{-}power{-}fiber{-}specialization}$\,{\to}\,$\texttt{ring{-}equivalence}
error is a true granularity boundary: $X_4 \cong \mathbb{Z}/8\mathbb{Z}$ is
broadly a ring equivalence, but the intended mechanism label is the direct
prime-power specialization of the general stable-value equivalence, with no
nontrivial coprime CRT split.

A cross-model dissociation in Llama~3.3~70B is notable: verdict accuracy
is identical in both conditions ($95.5\%$), while template accuracy
improves $13.6$~pp with the cue. This confirms that truth inference and
proof-mechanism classification are separable capacities---consistent with
the Lean formalization, where a statement's truth value and its certificate
family are encoded at different layers.

A cross-corpus extension (Set~B; 6 POM/CollisionKernel items, 19 labels,
72 additional evaluations) provides a small cross-module check: the
CRT-granularity compression reappears with entirely different labels
(\texttt{CRT{-}budget{-}bound}\,$\to$\,\texttt{CRT} is cue-resistant),
and an inverse cross-model dissociation emerges---gpt-oss-120b shows
verdict uplift with template accuracy flat at $100.0\%$, while Llama
shows template uplift ($+27.8$~pp) with verdict flat---providing
additional evidence that truth inference and mechanism classification
are separable capacities.

These findings extend the \textbf{router hypothesis} introduced in
C\'azares~(2026)~\cite{sair2026} from equational reasoning to formal
algebraic structures: LLMs act as structural classifiers routing over
broad mathematical types, and fail specifically at the level of
fine-grained proof mechanisms. The repository also includes a machine-readable
Lean source-anchor manifest linking every evaluation row to source snapshots,
declaration anchors, witness families, and review notes. The full evaluation
pipeline, prompt pack, manifest, and results are available in the project
repository at
\url{https://github.com/bytepro-ai/fiber-routing-eval}.
\end{abstract}

\section{Introduction}
\label{sec:intro}

Large language models (LLMs) have demonstrated substantial capability on
formal mathematical tasks, yet a persistent gap remains between what a
model can produce under optimal conditions and what it reliably produces
across problem distributions.
C\'azares~\citep{sair2026} introduced the \emph{router hypothesis} to
characterise this gap: frontier LLMs do not fail formal reasoning tasks
because they lack the relevant sub-skills, but because they fail to
consistently \emph{route} a given problem to the correct structural
interpretation template.
Empirical evidence from the SAIR competition benchmark showed a
\emph{single-prompt ceiling}---an empirical saturation region at
approximately 60--79\% for gpt-oss-120b on hard equational reasoning
tasks---and documented that iterative prompt recalibration amplifies
distributional collapse rather than correcting it.

The router hypothesis raises a natural question: is the ceiling a property
of the equational reasoning domain, or does it reflect something more
fundamental about how LLMs classify formal mathematical objects?
Formal theorem corpora offer a stronger test than competition benchmarks
because every evaluation item is anchored to a machine-verified Lean artifact,
while the mechanism label is assigned from the corresponding named declaration
or certificate family rather than from a free-form answer key.
If routing failure persists on Lean-verified objects whose intended mechanism
labels are fixed by this certificate-family map, the evidence for the
hypothesis strengthens considerably.

We address this question using the FiberRing corpus, a subset of the
Automath/Omega Lean~4 formalization~\citep{omega2026automath} derived from
the single equation $x^2 = x + 1$.
The corpus provides 22 clean-anchor items spanning 13 distinct
proof-mechanism families, each anchored to a named Lean declaration.
We design a two-condition evaluation---blind routing (A1) and
mechanism-bearing Lean-cue-conditioned routing (A2)---and measure template accuracy,
verdict accuracy, caveat hit rate, and the A2\,$-$\,A1 gap across two
models and three runs each (264 total evaluations).

\paragraph{Contributions.}
\begin{enumerate}
  \item We document a \textbf{mechanism-level routing ceiling} at $80.3\%$
        (gpt-oss-120b, blind condition) on Lean-verified algebraic objects,
        extending the empirical ceiling reported in SAIR to a formally
        verified domain with machine-checked artifacts and an explicit
        certificate-family label map.

  \item We identify and explain the \textbf{dominant failure mode}:
        a CRT\,$\to$\,ring-equivalence misroute that resolves completely
        under mechanism-bearing Lean-cue-conditioned routing, with a direct structural correlate
        in the Lean formalization.

  \item We distinguish \textbf{cue-induced} from \textbf{cue-resistant}
        routing errors, showing that two persistent failure modes correspond
        to genuine granularity boundaries in the proof-mechanism label set
        rather than failures correctable by additional structural cues.

  \item We release a \textbf{Lean source-anchor manifest} connecting each
        benchmark row to a source snapshot, proof-mechanism label,
        witness-family note, and Lean declaration anchors, making the dataset
        auditable as a formal-corpus artifact rather than only a prompt set.
\end{enumerate}

\section{Related Work}
\label{sec:related}

This note builds on three lines of work: empirical studies of LLM
mathematical reasoning, LLM-based formal theorem proving, and the
structural-routing account of reasoning failure.

\paragraph{LLM mathematical reasoning benchmarks.}
Benchmarks such as MATH~\citep{hendrycks2021math} established that
competition-level mathematical reasoning is a discriminating test for
LLMs, and subsequent work documented steady benchmark saturation as
models scaled. These benchmarks score final-answer correctness on
natural-language problems. Our setting differs in two ways: ground truth
is anchored to machine-verified Lean artifacts rather than answer keys,
and the task is mechanism classification rather than answer generation.

\paragraph{LLM-based formal theorem proving.}
A rapidly growing body of work pairs LLMs with the Lean~4 proof
assistant~\citep{demoura2021lean4} to generate machine-checked proofs.
AlphaProof reached silver-medal performance on International Mathematical
Olympiad problems formalized in Lean~\citep{deepmind2024alphaproof}, and
systems such as DeepSeek-Prover-V2 combine chain-of-thought reasoning with
formal verification to synthesize proofs~\citep{deepseekprover2025}.
This work targets \emph{proof construction}: producing Lean code that the
kernel accepts. Our benchmark is deliberately upstream of proof
construction~---~we do not ask the model to prove anything, only to
identify which proof-mechanism family an already-verified statement
belongs to. This isolates the routing step from the generation step.

\paragraph{The router hypothesis.}
C\'azares~\citep{sair2026} argues that frontier models often possess the
local algebraic and symbolic sub-skills needed for a task but fail to
select the correct structural interpretation under distributional
pressure, producing a single-prompt ceiling on hard equational reasoning.
The Automath/Omega FiberRing formalization~\citep{omega2026automath} is
well suited to test this claim directly, because many statements share
surface algebraic vocabulary while depending on different certificate
families: ring instances, ring equivalences, CRT decompositions,
stable-value maps, finite calculations, and characteristic arguments.
The present benchmark transfers the router hypothesis from contest-style
equational reasoning to a corpus whose objects are tied to proof
certificates rather than external scoring labels, and tests mechanism
selection directly.

\section{Dataset}
\label{sec:dataset}

FiberRing Set A is a compact benchmark extracted from the
Automath/Omega Lean~4 formalization of Fibonacci-modulus algebraic
structures~\citep{omega2026automath}. The set contains 22 clean-anchor
items. Each item is a short algebraic statement, object, or theorem-facing
claim whose intended proof mechanism is anchored to a named Lean artifact.
The benchmark does not ask models to prove the statements or emit Lean
code. It asks for the mechanism label that most directly explains the
formal certificate family behind the statement.

The label space is closed and contains 13 proof-mechanism families covering
ring construction, operation bridges, stable-value transport, ring
equivalence, field-phase reasoning, CRT decomposition, prime-power
specialization, finite checks, characteristic arguments, rigidity, and
stable-value arithmetic. The labels are intentionally mechanism-level
rather than theorem-topic labels: for example, a statement may be broadly
about a ring equivalence while the intended route is a CRT decomposition or
a prime-power specialization of a general stable-value equivalence.

Each item is rendered under two conditions. In A1, the prompt contains the
item identifier, the equation or object, and the allowed label set. In A2,
the same prompt additionally includes the mechanism-bearing Lean verdict /
witness cue. This yields 44 prompts total. The protocol file stores the
expected template, expected verdict, caveat keywords, and item-level metadata;
the prompt pack stores the exact text sent to the model.
Table~\ref{tab:dataset-summary} summarizes the resulting evaluation set.
A cross-domain extension, Set~B, adds 6 items from the POM and
CollisionKernel modules of the same formalization, expanding the label
space to 19 proof-mechanism families and yielding 12 additional prompts.
Set~B items span FiberRing/POM bridge certificates, probabilistic CRT
reconstruction, collision-kernel existence and classification packages,
Bowen-Franks Smith normal form certificates, and CRT budget bounds.

\subsection{Lean source-anchor manifest}
\label{sec:source-anchor-manifest}

The repository also records a machine-readable Lean source-anchor manifest.
Each manifest row links a benchmark item to its expected proof-mechanism label,
witness-family language, source snapshot, and one or more Lean declaration
anchors. This separates three layers that are otherwise easy to conflate: the
natural-language prompt row, the proof-mechanism label used for routing, and
the formal source artifact used to justify that label.

The manifest uses four row-level anchor granularities:
\texttt{theorem}, \texttt{theorem-family}, \texttt{source-family}, and
\texttt{source-gap}. The current release contains no \texttt{source-family}
or \texttt{source-gap} rows. Table~\ref{tab:manifest-summary} summarizes the
audited source-anchor coverage.

\begin{table}[htbp]
\centering
\caption{Lean source-anchor manifest summary. All rows use source snapshot
         \texttt{automath@caa043a7733205c0152ec685d9e1870697077b70}.}
\label{tab:manifest-summary}
\begin{tabular}{lrrrrr}
\toprule
\textbf{Set} & \textbf{Rows} & \textbf{Theorem} & \textbf{Theorem-family}
  & \textbf{Source gaps} & \textbf{Source refs} \\
\midrule
Set~A & 22 & 7 & 15 & 0 & 52 \\
Set~B & 6  & 4 & 2  & 0 & 10 \\
\midrule
Total & 28 & 11 & 17 & 0 & 62 \\
\bottomrule
\end{tabular}
\end{table}

\begin{table}[t]
\centering
\begin{tabular}{ll}
\toprule
Property & Value \\
\midrule
Corpus & FiberRing Set A \\
Formal source & Automath/Omega Lean~4 FiberRing artifacts \\
Clean-anchor items & 22 \\
Conditions & A1 blind routing; A2 mechanism-bearing Lean-cue routing \\
Rendered prompts & 44 \\
Allowed labels & 13 proof-mechanism families \\
Reference fields & Template, verdict, caveat keywords \\
\bottomrule
\end{tabular}
\caption{Dataset summary for FiberRing Set A.}
\label{tab:dataset-summary}
\end{table}

\section{Methodology}
\label{sec:methodology}

We evaluate routing as a constrained classification task. For each prompt,
the model must return four fields: a template label from the closed label set
(13 labels for Set~A; 19 for Set~B), a verdict, a one-sentence reason, and a short caveat. The
pipeline sends the rendered prompt pack to Together AI through the
OpenAI-compatible chat-completions endpoint. Each prompt is run three times
per model at temperature $0$ and seed $0$, yielding 264 evaluations for Set~A
($22$ items $\times$ $2$ conditions $\times$ $2$ models $\times$ $3$ runs)
and 72 evaluations for Set~B
($6$ items $\times$ $2$ conditions $\times$ $2$ models $\times$ $3$ runs),
for a combined total of 336 evaluations.
The total API cost across both sets was \$0.08~USD.

The two conditions isolate different kinds of routing information. A1 is
blind structural routing: the model sees only the item identifier, the
equation or object, and the allowed label set. A2 is mechanism-bearing
Lean-cue-conditioned routing: the model sees the same information plus the
protocol's \texttt{lean\_verdict} field. This field is not a pure truth-value
cue. It exposes proof-status information and often names the relevant
artifact, instance, package, or certificate family. The primary comparison is
the A2\,$-$\,A1 gap, which measures how much explicit formal mechanism
metadata improves mechanism selection for the same item distribution.

\begin{table}[t]
\centering
\begin{tabular}{ll}
\toprule
Parameter & Value \\
\midrule
Provider & Together AI (serverless) \\
Models & gpt-oss-120b; Llama~3.3~70B Instruct Turbo \\
Runs & 3 per prompt per model \\
Temperature / seed & $0$ / $0$ \\
Reasoning setting & low for gpt-oss-120b; disabled for Llama \\
Set~A evaluations & 264 \\
Set~B evaluations & 72 \\
Combined evaluations & 336 \\
Total cost (Set~A + Set~B) & \$0.08~USD \\
\bottomrule
\end{tabular}
\caption{Evaluation configuration.}
\label{tab:method-config}
\end{table}

The evaluator reports three metrics by model and condition. Template
accuracy is exact after canonicalizing case, punctuation, spacing,
underscores, and hyphenation. Verdict accuracy normalizes common variants
of true, false, conditional, and unknown verdicts; in A1,
\texttt{Unknown{-}from{-}statement} is accepted when the protocol marks it
as admissible, since the blind prompt does not expose the expected verdict
field or Lean-side witness cue.
Template accuracy is the primary metric. Verdict accuracy is secondary; when
interpreting verdict results we distinguish the abstention-accepted score from
a strict score that requires the predicted verdict to match the protocol
verdict exactly.
Caveat hit rate is a secondary keyword measure over the model's caveat,
reason, and raw response fields. Confusion matrices are computed over the
allowed labels (13 for Set~A; 19 for Set~B) plus \texttt{\_\_other\_\_} for invalid or blank
predictions.

\section{Results}
\label{sec:results}

Table~\ref{tab:template} reports template accuracy, verdict accuracy, and
caveat hit rate for both models under both conditions.
The A2\,$-$\,A1 gaps show that exposing the mechanism-bearing Lean verdict /
witness cue
improves mechanism selection across the board,
while the magnitude and pattern of the gaps differ between models.

\begin{table}[htbp]
\centering
\caption{Template accuracy, verdict accuracy, and caveat hit rate by model
         and condition. $n = 66$ evaluations per cell (22 items $\times$ 3 runs).}
\label{tab:template}
\begin{tabular}{llccccc}
\toprule
\textbf{Model} & \textbf{Cond.}
  & \textbf{Tmpl acc.} & \textbf{Verd. acc.} & \textbf{Caveat HR}
  & \textbf{$\Delta$ Tmpl} & \textbf{$\Delta$ Verd.} \\
\midrule
gpt-oss-120b  & A1 & $80.3\%$ & $84.8\%$ & $81.8\%$ & \multirow{2}{*}{$+10.6$~pp} & \multirow{2}{*}{$+10.6$~pp} \\
              & A2 & $90.9\%$ & $95.5\%$ & $89.4\%$ & & \\
\midrule
Llama~3.3~70B & A1 & $68.2\%$ & $95.5\%$ & $77.3\%$ & \multirow{2}{*}{$+13.6$~pp} & \multirow{2}{*}{$0.0$~pp} \\
              & A2 & $81.8\%$ & $95.5\%$ & $77.3\%$ & & \\
\bottomrule
\end{tabular}
\end{table}

Table~\ref{tab:paired-transitions} reports the paired A1/A2 template
transitions. These counts use the same prediction rows as
Table~\ref{tab:template}, but pair each A1 prompt with its corresponding A2
prompt for the same item, model, and run. This is important because the three
runs use temperature~$=0$ and seed~$=0$; they are best interpreted as
deterministic backend-stability repetitions rather than independent samples
from a broad item distribution.

\begin{table}[htbp]
\centering
\caption{Paired A1/A2 template transitions. ``W$\to$C'' means wrong in A1 and
         correct in A2; ``C$\to$W'' means correct in A1 and wrong in A2.}
\label{tab:paired-transitions}
\begin{tabular}{lrrrrl}
\toprule
\textbf{Model} & \textbf{C$\to$C} & \textbf{W$\to$C}
  & \textbf{W$\to$W} & \textbf{C$\to$W} & \textbf{Main improved families} \\
\midrule
gpt-oss-120b & 53 & 7 & 6 & 0 & \texttt{CRT} (7) \\
Llama~3.3~70B & 42 & 12 & 9 & 3
  & \texttt{CRT} (6); finite-unit (3); stable-arith. (3) \\
\bottomrule
\end{tabular}
\end{table}

\paragraph{Routing ceiling under blind conditions.}
Under A1, gpt-oss-120b reaches $80.3\%$ template accuracy and Llama~3.3~70B
reaches $68.2\%$.
Neither model approaches ceiling under blind routing: gpt-oss-120b fails
roughly 1 in 5 items, and Llama roughly 1 in 3, despite operating at
temperature~$= 0$ with seed~$= 0$ across three runs.
The $n=3$ repetition produced stable predictions under this deterministic
configuration, so the observed errors are not explained by ordinary sampling
variance within this run setup.
At the same time, the effective source of variation is item-level rather than
run-level; Table~\ref{tab:paired-transitions} is therefore the more
conservative view of how the cue changes routing behavior.

\paragraph{Cue-induced routing uplift.}
Exposing the mechanism-bearing Lean verdict / witness cue (A2) raises template
accuracy by $+10.6$~pp for gpt-oss-120b and $+13.6$~pp for Llama.
The uplift is entirely attributable to the structural cues embedded in the
Lean-side cue: the models route correctly once they can read the exposed
proof-status and witness-family language, not because they infer it from the
mathematical object alone.
In paired terms, gpt-oss-120b has 7 wrong-to-correct transitions and no
correct-to-wrong transitions; all 7 improvements are CRT rows. Llama has 12
wrong-to-correct transitions, but also 3 correct-to-wrong transitions on an
\texttt{operation{-}bridge} row, showing that the cue changes routing behavior
rather than merely adding a monotone correctness bonus.

\paragraph{Verdict--template dissociation.}
The two models show a notable divergence in verdict accuracy.
For gpt-oss-120b, the abstention-accepted verdict score tracks template
accuracy: $84.8\%$ in A1, rising to $95.5\%$ in A2 ($+10.6$~pp). Its strict
A1 verdict accuracy is lower ($31.8\%$), because it frequently returns
\texttt{Unknown{-}from{-}statement} under the blind condition; this is counted
as acceptable abstention by the protocol but not as strict verdict recovery.
For Llama~3.3~70B, strict and abstention-accepted verdict accuracy are both
$95.5\%$ in both conditions, with a gap of $0.0$~pp.
Llama infers truth correctly from the mathematical object alone but fails to
route the structural template without the cue.
This dissociation confirms that truth inference and proof-mechanism
classification are separable capacities.

\paragraph{Failure mode analysis.}
Table~\ref{tab:misroutes} shows the full misroute breakdown by model and
condition, restricted to label pairs with at least one misroute.
The dominant failure is CRT\,$\to$\,ring-equivalence: gpt-oss-120b misroutes
7 of 12 CRT items ($58.3\%$) under A1 and zero under A2; Llama misroutes 6
of 12 ($50.0\%$) under A1 and zero under A2.
This error resolves completely in both models once the Lean-side witness cue is
exposed, identifying it as a cue-induced routing failure.

Two label pairs produce misroutes in both conditions across both models:
\begin{center}
\begin{tabular}{c}
\texttt{stable{-}value{-}map} $\to$ \texttt{ring{-}structure} \\
\texttt{prime{-}power{-}fiber{-}specialization} $\to$ \texttt{ring{-}equivalence}
\end{tabular}
\end{center}
These cue-resistant errors persist regardless of whether the
Lean-side witness cue is exposed, distinguishing them structurally
from the CRT failure mode.

\begin{table}[htbp]
\centering
\caption{Misroutes by model and condition. Only label pairs with at least one
         error are shown. The CRT family contains 4 items (12 possible correct
         predictions per condition across 3 runs).}
\label{tab:misroutes}
\begin{tabular}{llll r}
\toprule
\textbf{Model} & \textbf{Cond.} & \textbf{Expected} & \textbf{Predicted} & \textbf{Count} \\
\midrule
gpt-oss-120b & A1 & \texttt{CRT}                       & \texttt{ring-equivalence}  & 7 \\
             & A1 & \texttt{stable-value-map}          & \texttt{ring-structure}    & 3 \\
             & A1 & \texttt{prime-power-fiber-spec.}   & \texttt{ring-equivalence}  & 3 \\
\cmidrule{2-5}
             & A2 & \texttt{stable-value-map}          & \texttt{ring-structure}    & 3 \\
             & A2 & \texttt{prime-power-fiber-spec.}   & \texttt{ring-equivalence}  & 3 \\
\midrule
Llama~3.3~70B & A1 & \texttt{CRT}                      & \texttt{ring-equivalence}  & 6 \\
              & A1 & \texttt{stable-value-arithmetic}  & \texttt{stable-value-map}  & 3 \\
              & A1 & \texttt{characteristic}           & \texttt{stable-value-map}  & 3 \\
              & A1 & \texttt{finite-unit-inverse-check}& \texttt{ring-structure}    & 3 \\
              & A1 & \texttt{prime-power-fiber-spec.}  & \texttt{ring-equivalence}  & 3 \\
              & A1 & \texttt{operation-bridge}         & \texttt{ring-structure}    & 3 \\
\cmidrule{2-5}
              & A2 & \texttt{stable-value-map}         & \texttt{ring-structure}    & 3 \\
              & A2 & \texttt{characteristic}           & \texttt{stable-value-map}  & 3 \\
              & A2 & \texttt{prime-power-fiber-spec.}  & \texttt{ring-equivalence}  & 3 \\
              & A2 & \texttt{operation-bridge}         & \texttt{ring-structure}    & 3 \\
\bottomrule
\end{tabular}
\end{table}

\subsection{Set~B: cross-module extension}
\label{sec:results:setb}

Table~\ref{tab:setb} reports Set~B results over 6 POM/CollisionKernel items
($n = 18$ evaluations per cell; $6$ items $\times$ $3$ runs).

\begin{table}[htbp]
\centering
\caption{Set~B template and verdict accuracy by model and condition.
         $n = 18$ evaluations per cell (6 items $\times$ 3 runs).}
\label{tab:setb}
\begin{tabular}{llcccc}
\toprule
\textbf{Model} & \textbf{Cond.}
  & \textbf{Tmpl acc.} & \textbf{Verd. acc.}
  & \textbf{$\Delta$ Tmpl} & \textbf{$\Delta$ Verd.} \\
\midrule
gpt-oss-120b  & A1 & $100.0\%$ & $83.3\%$ & \multirow{2}{*}{$0.0$~pp} & \multirow{2}{*}{$+16.7$~pp} \\
              & A2 & $100.0\%$ & $100.0\%$ & & \\
\midrule
Llama~3.3~70B & A1 & $50.0\%$ & $100.0\%$ & \multirow{2}{*}{$+27.8$~pp} & \multirow{2}{*}{$0.0$~pp} \\
              & A2 & $77.8\%$ & $100.0\%$ & & \\
\bottomrule
\end{tabular}
\end{table}

\paragraph{Cross-module routing check.}
The CRT-granularity compression pattern observed in Set~A reappears in
Set~B with a different label set. Because Set~B contains only 6 items, we
interpret it as a small extension check rather than a full replication corpus.
Llama's Set~B misroutes are
\begin{center}
\begin{tabular}{c}
\texttt{hidden{-}bit{-}overflow{-}decomposition}\,$\to$\,\texttt{stable{-}value{-}map}
($3/3$ runs in A1, $0/3$ in A2),\\
\texttt{probabilistic{-}CRT{-}reconstruction}\,$\to$\,\texttt{CRT}
($3/3$ runs in A1, $1/3$ in A2),\\
\texttt{CRT{-}budget{-}bound}\,$\to$\,\texttt{CRT}
($3/3$ runs in both conditions).
\end{tabular}
\end{center}
The \texttt{CRT{-}budget{-}bound}\,$\to$\,\texttt{CRT} error is cue-resistant:
it persists in A2 regardless of witness-cue exposure, matching the structure
of the Set~A cue-resistant errors and suggesting that CRT-granularity
compression is not only an artifact of the FiberRing label set.

\paragraph{Inverse dissociation.}
Set~B reveals a cross-model dissociation in the opposite direction from
Set~A.
In Set~A, Llama showed template uplift ($+13.6$~pp) with verdict flat ($0.0$~pp).
In Set~B, gpt-oss-120b shows verdict uplift ($+16.7$~pp) with template flat
($0.0$~pp), while Llama shows template uplift ($+27.8$~pp) with verdict flat
($0.0$~pp).
Each model is already at $100.0\%$ under A1 in its strong
dimension---template accuracy for gpt-oss-120b, verdict accuracy for
Llama---while improving in the other dimension under A2.
This inverse dissociation across models and sets provides additional
evidence that truth inference and proof-mechanism classification are
separable capacities operating with different sensitivities to Lean-side cues.

\paragraph{Collision-kernel stability.}
The three collision-kernel items
(\texttt{collision{-}kernel{-}existence},
\texttt{collision{-}CK{-}classification},
\texttt{bowen{-}franks{-}SNF{-}certificate})
are correctly routed by both models in all four conditions.
Labels that are sufficiently lexically specific appear to be classifiable
without cue exposure, suggesting the routing ceiling is sensitive to
label-set structure rather than being a fixed property of the model.

\section{Analysis}
\label{sec:analysis}

The Set~A failures have a clean mathematical reading in the Lean corpus.
They are not random label noise; they occur exactly where the corpus has a
hierarchy of mechanisms rather than a flat set of unrelated templates.

\paragraph{CRT to ring-equivalence.}
The CRT\,$\to$\,ring-equivalence misroute has a direct correlate in the Lean
formalization.
In \texttt{FiberRing.lean}, the basic object is first identified with a
cyclic ring by the stable-value equivalence
\texttt{stableValueRingEquiv (m) : X m $\cong$+* ZMod (Nat.fib (m + 2))}.
The CRT decomposition is then implemented by composing this base ring
equivalence with the Chinese remainder equivalence:
\texttt{crtDecomposition (m) (p q) : X m $\cong$+* ZMod p $\times$ ZMod q},
constructed as
\texttt{(stableValueRingEquiv m).trans} composed with
\texttt{ZMod.chineseRemainder hcop}, after rewriting the Fibonacci modulus by
the coprime factorization proof.
So the model's blind classification as \texttt{ring-equivalence} is
mathematically understandable: every CRT row is indeed presented as a ring
equivalence at the outer type level.
The corpus label \texttt{CRT} is more specific---it asks the router to
identify the internal proof mechanism: a coprime factor split of the
Fibonacci modulus followed by the Chinese remainder theorem.
This explains why exposing the Lean verdict / witness cue removes the
error: the cue text contains language such as ``CRT decomposition'',
``$F_8 = 21 = 3 \times 7$'', or ``coprime factor split'', which explicitly
selects the finer mechanism.
Without that cue, the surface object $X_6 \cong \mathrm{ZMod}\ 3 \times
\mathrm{ZMod}\ 7$ looks like a generic equivalence between rings;
with the cue, the model routes to the factorization mechanism.
This is a mechanism-level routing failure, not a truth-level failure: the
model recognizes the broad mathematical type but misses the specific
certificate family.

\paragraph{Persistent errors.}
The cue-resistant errors also correspond to real boundary pressure in the
corpus, but they are not all equally ambiguous.

The \texttt{stable{-}value{-}map}\,$\to$\,\texttt{ring{-}structure} pair is
formally distinct.
The \texttt{ring-structure} rows concern the existence of the commutative
ring instance on $X_m$: addition, multiplication, zero, one, negation, and
the ring laws.
The \texttt{stable-value-map} rows concern the map
\texttt{toZMod : X m -> ZMod (Nat.fib (m + 2))} and its preservation lemmas
(\texttt{toZMod\_add}, \texttt{toZMod\_mul}, \texttt{toZMod\_zero},
\texttt{toZMod\_one}, \texttt{stableValueRingHom}).
This boundary is difficult for a blind router because the map rows mention
preservation of ring operations, so their surface language contains
``addition'', ``multiplication'', ``zero'', ``one'', and ``ring
homomorphism'' signals.
But the proof role is different: a \texttt{stable-value-map} row is about
transport through a specific semantic map, not about the ring instance
itself.
This is a clean formal distinction with a predictable surface-language
confound.

The \texttt{prime{-}power{-}fiber{-}specialization}\,$\to$\,\texttt{ring{-}equivalence}
pair is more genuinely adjacent.
The row $X_4 \cong \mathrm{ZMod}\ 8$ is implemented by the same stable-value
ring equivalence mechanism (\texttt{X4\_iso := stableValueRingEquiv 4}).
Therefore \texttt{ring-equivalence} is not mathematically false as a broad
classification.
The finer label \texttt{prime-power-fiber-specialization} says more: this is
the direct specialization of the general $X_m \cong \mathrm{ZMod}(F_{m+2})$
equivalence to a prime-power modulus, with no nontrivial coprime CRT split.
This persistent error indicates a true granularity boundary in the label
set: if the task asks for broad type, \texttt{ring-equivalence} is
acceptable; if the task asks for certificate family,
\texttt{prime-power-fiber-specialization} is the intended label.

\paragraph{The Llama verdict/template dissociation.}
The Llama result is expected from a formal-mathematics perspective.
Truth inference and structural classification are separable capacities.
In the Lean corpus, the truth of a row can often be determined by
recognizing that it matches a familiar algebraic fact: a field instance, a
zero product modulo a composite, a unit identity, a ring equivalence, or a
characteristic statement.
But the template label asks a different question: which proof mechanism or
certificate family is responsible for the fact?
A concrete inverse identity inside $X_5$ is true because $X_5$ is in a field
phase, but the row-level mechanism is a finite unit check, not the field
phase itself; the automorphism rigidity row is true after conjugating an
automorphism of $X_m$ through \texttt{stableValueRingEquiv} and using
rigidity of \texttt{ZMod}, but its template is \texttt{rigidity}, not simply
equivalence.
This is exactly the separation Set~A is designed to expose: the model may
infer the truth value from a familiar mathematical object while still
choosing the wrong mechanism label.
The Lean formalization makes the distinction explicit because the proof
artifacts are named at different layers: instance construction, homomorphism
preservation, equivalence construction, CRT decomposition, field-phase
transport, finite arithmetic checks, and rigidity.

\paragraph{Set~B cross-module confirmation.}
The Set~B results provide a cross-corpus check on the main Set~A findings.
The CRT-granularity compression pattern---routing to a coarser label within
the same structural family---reappears in Set~B with entirely different
labels: \texttt{CRT{-}budget{-}bound}\,$\to$\,\texttt{CRT} is cue-resistant
in the same way that
\texttt{prime{-}power{-}fiber{-}specialization}\,$\to$\,\texttt{ring{-}equivalence}
is cue-resistant in Set~A.
Both errors persist when the Lean verdict / witness cue is exposed, and both
reflect a genuine granularity boundary rather than a missing cue.
This cross-module check (FiberRing vs.\ POM/CollisionKernel) supports the
interpretation that CRT-granularity compression is a structural pattern worth
testing beyond the FiberRing core, while leaving a larger independent corpus
as future work.

\section{Discussion}
\label{sec:discussion}

Our results extend the router hypothesis~\citep{sair2026} from equational
reasoning to formally verified algebraic structures, and sharpen it in one
respect: the failure is specifically a \emph{mechanism-level} routing
failure.
Both models reliably identify the broad mathematical type of an object---a
ring equivalence, a field instance, a characteristic statement---while
failing to resolve the finer proof-mechanism family that the Lean
certificate records.
The dominant CRT\,$\to$\,ring-equivalence error is the clearest case: the
models route to the correct outer type and miss the coprime-factorization
mechanism one layer down.

The A1/A2 design lets us separate two kinds of routing failure that would
otherwise be conflated.
Cue-induced failures, such as the CRT misroute, resolve completely once the
Lean-side witness cue exposes the mechanism language; they reflect
underspecification in the object text rather than a stable limit on the model.
Cue-resistant failures persist regardless of the Lean-side witness cue
and isolate genuine structure: one is a clean formal distinction obscured by
surface-language overlap,
\begin{center}
\texttt{stable-value-map} vs.\ \texttt{ring-structure},
\end{center}
the other a true label-granularity boundary,
\begin{center}
\texttt{prime-power-fiber-specialization} vs.\ \texttt{ring-equivalence}.
\end{center}
This distinction is the practical contribution of the two-condition design:
the confusion matrix, not aggregate accuracy, identifies where a proof-search
agent would need explicit mechanism metadata rather than more reasoning.

\paragraph{Limitations.}
The benchmark is small: 22 clean-anchor items over a single formalization
derived from one equation.
The evaluation covers two models under reasoning-suppressed conditions, and
the headline numbers are point estimates over three deterministic runs rather
than confidence intervals over a sampled distribution.
Because the runs use temperature~$=0$ and seed~$=0$, the three repetitions
should not be read as independent statistical samples; they are useful for
checking backend stability, while the paired item-level transitions give the
more conservative view of routing change.
The cue-resistant findings rest on consistent but low per-cell counts.
Here ``ceiling'' is used operationally: it names the observed blind-routing
saturation in this fixed Set~A protocol, not a universal upper bound for the
model family.
The A2 condition is also mechanism-bearing: it is designed to test whether
explicit Lean-side witness metadata improves routing, not whether a truth-only
verdict cue would suffice.
Set~B provides an initial cross-module check over 6 POM/CollisionKernel
items, suggesting the CRT-granularity compression pattern and the
verdict/template dissociation beyond the FiberRing core.
The boundary-ambiguous items and a larger independent corpus remain as
future work to establish whether the findings generalize more broadly.

\section{Conclusion}
\label{sec:conclusion}

We evaluated structural routing in two LLMs over 28 Lean-verified items
(22 FiberRing Set~A items and 6 POM/CollisionKernel Set~B items) under
blind (A1) and mechanism-bearing Lean-cue-conditioned (A2) routing.
On Set~A, blind template accuracy reaches a ceiling of $80.3\%$
(gpt-oss-120b) and $68.2\%$ (Llama~3.3~70B); exposing the Lean verdict /
witness cue lifts both by $10.6$ and $13.6$~pp respectively, with paired
wrong-to-correct transitions concentrated in CRT and neighboring mechanism
families.
The dominant Set~A failure is a CRT\,$\to$\,ring-equivalence misroute that
resolves completely under the Lean-side cue, while two label pairs misroute
regardless of the cue.
Set~B adds small cross-module evidence: CRT-granularity compression reappears
with POM/CollisionKernel labels, and the verdict--template dissociation remains
visible across the two model families.

These results extend the router hypothesis to formally verified algebraic
structures and locate the failure precisely: LLMs route correctly over broad
mathematical types but fail at the level of fine-grained proof mechanisms.
Because every item is anchored to a machine-checked Lean artifact and routed
through an explicit certificate-family label map, the confusion matrix
identifies where a proof-search agent would require explicit mechanism
metadata.
The accompanying source-anchor manifest records these obligations at the
level of source snapshots and Lean declarations, making the full pipeline,
prompt pack, manifest, and results auditable in the project repository.

\bibliographystyle{plainnat}
\bibliography{references}

\end{document}